# Towards a Better Understanding of CAR, CDR, CADR and the Others

Thomas W. Lynch
Researcher at Birkbeck College University of London
Thomas.Lynch@reasoningtechnology.com
2015-07-20

## Introduction

This article introduces a method for teaching the *CAR* and *CDR* extensions.

In programming, a *container* has multiple cells for holding other objects. We use an access language to specify a cell and say what action to take when it is found, such as returning the object in the cell.

As an example, in FORTRAN, the primary container is the array, and there is a simple sub-language inside of FORTRAN used for accessing an element within an array. Namely,

>   *symbol* ( *index* )

Here, the parenthesis are literal punctuation, the *symbol* will represent the base address for an array, and the *index* will be multiplied by the element size and added to the base address for finding an element.

The primary container type in LISP is the list. The elements of a list may also be lists, so the primary container type is actually that of a tree. LISP, like FORTRAN, can be thought of as having a sub-language for accessing elements from objects of the primary container type. This sub-language has at its root the functions *CAR* and *CDR*. Before describing that sub-language lets examine the genesis, of these function names.

## *Origin of the Function Names CAR and CDR*

John McCarthy describes the history of *CAR* and *CDR* in [1], but after reading this, many of us are still confused. After all, what is a *decrement register* and how can it be so important to LISP?

Scientific computing uses fixed-length integers and floating-point values held in rigid containers known as arrays. In scientific computing, an array is analogous to a vector, and a two-dimensional array analogous to a matrix. Matrix operations are of tantamount importance in scientific computing, and during a matrix multiplication, a program typically walks through matrices element by element. As a consequence of using a fixed-length number representation, such a walk starts at a base element address, then progresses by successively adding the constant element-length.

AI computation is symbol-based rather than number based. Containers are more commonly variable-length property lists with variable-length elements rather than fixed-length vectors of constant element-length. LISP lists are built from a series of connected nodes known as *cons cells*. Each cons cell is an address pair, where the first address locates the list item, and the second address locates the next cell. To walk through the list, we start with a base address that locates the first cell. Then, to take successive steps, instead of adding an element-length to the base register at each step, we instead completely replace the contents of the base register with the next-cell-address from the current cell.

In LISP the functions CAR and CDR operate on cons cells. The function CAR first extracts from the cell the address that locates the list item, and then it returns the item. The function CDR extracts from the cons cell the next cell address, and returns it.

LISP was developed on a mass-produced computer that was sold to the public, the IBM 704. It was a mature design, not a research architecture. Thanks to an MIT instruction manual on programming the IBM 704, one co-authored by John McCarthy, we can get an idea about the technical language used when programming the 704 at the time LISP was developed [2]. MIT's programming manual for the 704 is substantially the same as IBM's [3], so the MIT programmers were using standard nomenclature. These manuals would be used by scientific programmers as well as those in the AI lab, so it is not surprising that there is an entire chapter dedicated to matrix addressing, entitled: "Indexing: Counting and Address Modification."

At the time LISP was developed in the 1950s, customers typically wanted scientific or business computers. The techniques of AI were largely not invented yet. Hence, the IBM 704 architecture had features for scientific computing (including floating-point arithmetic and addressing modes for walking arrays) and features for business computing (such as BCD number support), but no special features intended for AI applications.

The 704 organized arrays by placing the first element at the highest address and the last element at the lowest address. Hence, a programmer would decrement the base address to find successive elements rather than increment it. The base address would change depending on the location of the array, while the decrement value, which held the element length, would remain constant.

The 704 had two formats for instructions, *A* and *B*. Instruction format *A* had two 15-bit fields with 6 bits for the op code. The field layout was 3-15-3-15, so the two 15-bit fields were not directly adjacent. One 15-bit field was designated to hold an element address, while the other 15-bit field was designated to hold a decrement integer. Instruction format A was also used for data registers that held array base address / decrement integer pairs, both on the processor and in primary storage (system memory).

The 704 had three index registers to support complex addressing modes, including base offset addressing, automatic decrementing of indexes by a decrement value, and branching when index registers reached given values. These addressing modes were very useful for matrix multiplies, but of little help for traversing lists. The index registers were called A, B, and C in the IBM manual, and IR1,

IR2, and IR4 in the MIT manual.  (Index register three was not missing; rather it was the case that the index register specifiers were 'one hot' encoded so that multiple registers could be specified simultaneously.)

The 704 was an accumulator machine, with the accumulator affectionately called Ac. There was a dedicated processor register for holding a quotient or providing a multiplier, called MQ. There was one additional processor register that acted to buffer memory, called simply the *storage register*, or SR. The registers in system memory were referred to by numbers, or (one would assume) by variable names set from assembly mnemonics created by the programmer.

The 704's central processor had *no register named the "address register"* nor one named *"decrement register,"* either explicitly or in descriptions. Rather, the registers used for addressing or decrementing were called *index* registers.

The programming manuals first speak of the "two 15 bit fields" of instruction format A, but shortly after refer to these as the, "address part"  and "decrement part", apparently making a distinction between formatting and content.

Because memory had 36-bit words, while addresses were only 15-bits, the situation begged for some sort of packing scheme. *Machine Instructions were provided for this.*    The instruction LXA would load the contents of the address field from a word in memory, while LXD would load from the decrement field. The mnemonics table in the IBM manual describes LXD simply as "load index from decrement." Instructions that moved data were called either *load*, *store*, or *place* and each variation had a mnemonic that started with the letter L, S, or P, respectively.

Thus the LXA and LXD assembly instructions are semantically closest to the LISP functions CAR and CDR, but as we can see, their names are completely different.  *There were no assembly instructions in the 704 called CAR, or CDR, and this is not even close to being consistent with the naming convention*.

Using LXA and LXD  did not trigger the complex addressing modes, hence nothing prevented a programmer from, say, putting an address value into a decrement field.  Hence the LISP guys found it useful to use instruction format A registers to hold list nodes. They then used the decrement part for the next node address, and the element address part for the data address.  *Thus their  decrement part would never participate in the offset decrementing of an array element address.  The syntax 'decrement' has no semantic value for LISP.*

The IBM programmer's manual used the function capital letter 'C' as a shorthand to mean "contents of". As examples, "C(100)" would meant  "the contents of register 100", and "C(MQ)"  the "contents of the MQ register," etc. This was not reflected in the syntax of the assembly language (although it would be incorporated in later assembly languages). *The text of the manuals never refers to the "contents of an address register" or "contents of a decrement register," as such registers simply did not exist.* Rather, the manuals consistently use the nomenclature of the form, "the address part of C(*reg*)" to be read as the "the address part of the contents of the *reg* register," where *reg* is a memory address or the name of a processor register. Note, if we were to make a LISP function through abbreviation of "address part of C(reg)," it would be "APC(*reg*)". It would seem natural to assume the "Contents of", leading to, "AP(*reg*)"

It is stated in [1 – see node2.html] that CAR stands for "Contents of the Address part of Register." This is a permutation of the phrasing used in the manuals. Here 'part' must be used to mean 'field' rather than contents. The full abbreviations would be CAPR and CDPR. However when the "part" is dropped, the acronym becomes ambiguous. Recall that there were no registers known as *address* or *decrement* registers. Rather, there were two different parts to be distinguished. Without ambiguity, and with the 'contents of' permutation, these would shorten to "CAP(*reg*)" and "CDP(*reg*)" i.e. "Contents of Address Part" and "Contents of Data Part." So the LISP function names don't seem to make sense.

But we must ask a deeper question, why set the names of high level language functions, technically misnomers or not, to machine parts with differing and lower level purposes - instead of naming them for what they do? LISP did not even use the contents of decrement part of a register to do a decrement. Furthermore, if LISP was a success, it would be ported to other machines. Steve Russel explains that indeed there was regret, and soon after, they tried to convince students to use *first* and *rest* instead, "but it was too late" [4].

Let me suggest an explanation. Perhaps Steve and others noticed that list walking was much like array walking, with only a simple variation, that of replacing an address instead of incrementing it. There is some beauty in this, and it presages the later arrival of the iterator concept. Perhaps in noticing this the implementation detail was kept in mind and then manifested in the LISP language.

In general there is nothing wrong with having short cool-sounding but nonsensical terms for important new and oft used functions. For one thing, it removes the problem of aliasing against function names the user would like to have. *First*, for example, might have been a well-named function with a different purpose in many user contexts. And even when it doesn't alias, it *looks* like something that would belong to a user rather than a programming language. (Although, if you program for the auto industry, *CAR* might not be so great, either ;-) Another good thing about *CAR* and *CDR* as terms is that they can be extended into an access language to create functions like (CAR (CDR lst)) → (CADR lst) [5]. We explore this sort of composition in the remainder of this paper.

## The Elegant LISP Tree Access Language

I have heard that history can be rewritten, especially for the young. As an example, the */usr* directory in Unix is no longer the fumble-fingered user directory, one that evolved to have little to do with the user, but is now the grand "Unix System Resource." Perhaps a similar approach can be taken to make *CAR* and *CDR* more *elegant*.

The LISP tree access language consists of only four statements signified by single letters. These statements are then placed in sequence to create an access program. The four statements are as follows:

| a | access | accesses the head cell |
| c | complete | finished processing |
| d | drop | drop the head cell |
| r | run | run the program |

Tree access programs are written from right to left. This ensures the statements in the access program occur in the same scan order as if they were broken into individual *CAR* and *CDR* function applications. It also enables us to place an access program to the left of a list and have the active statement appear just next to the list head it is working on, as shown in the example below. The *run* and *complete* capstones help us distinguish access programs from symbols in our LISP program. Hence, you know in advance that if you use a symbol that starts with *c* and ends in *r*, and has only a combination *a* and *d* in between, you have aliased against an access program.

Here is an example, let's run *cadadr* on the tree '(0 (1 2 3) 4 5):

| 1. | cadad | '(0 (1 2 3) 4 5) | drop → '((1 2 3) 4 5) |
| 2. | cada | '( (1 2 3) 4 5) | access → '(1 2 3) |
| 3. | cad | '(1 2 3) | drop → '(2 3) |
| 4. | ca | '(2 3) | access → 2 |
| 5. | c | 2 | complete |

Now let's ask Common Lisp just to be sure ;-)

* (cadadr '(0 (1 2 3) 4 5))

In step 1 of this example, the first statement of our program says to drop the head of the list. We lose the zero. The new list is ('(123) 4 5).

Now in step 2, the next statement of our program says to access the head of the list. The head of the list is '(1 2 3).

The next statement of our program says to drop the head of the list. After dropping the 1, the list is now '(2 3).

The last statement of our program says to access the head of the list. This is the number 2. We then complete and return the 2.

## Potential Improvements

Suppose, when walking off the end of a tree, we instead returned a symbol such as 'EOT (end of tree). Notice that if we did this, then we wouldn't be able to use said symbol as a value within a tree. Yet, the very fact that we have introduced the symbol means we will want to put it somewhere, play with it, talk about it, log it, all while the tree is our primary container type. It follows that, in general, we cannot have a symbol to represent walking off the end of a tree, but must appeal to a higher level structure, such as an exception, continuation, control structure, or multiple return values – perhaps one being a primary return value and the second being an error code.

I suspect it would be easier to read the access programs if they were written from left to right. Accordingly, *CAR* and *CDR* would turn into *rac* and *rdc*. The example *cadadr* would become *rdadac*.

We might want to change the capstones, perhaps by using an operator to begin and a space at the end, so rdadac would become something like, *dada, "drop, access, drop, access." It is a road map through the tree.

Integer repeat counts would be helpful. As an example, *dddddda, a program that accesses the 6th element of a list, becomes *5da.

Another possible improvement would be to separate the location of a cell in the tree from operation on that cell or tree. Accordingly, *5n would locate the $6^{th}$ element. Here, I use *n* for *next*. Once it is located, we might take the prefix of the list to that point, the suffix from that point, destructively write the element, or return the element.

    (*5ns my-list)) ; returns a list consisting of elements 6 through last
    (*5nw! my-list 'apple) ; writes the $6^{th}$ cell
    (*5nr my-list);  reads the $5^{th}$ cell,  i.e. returns the contents of the $6^{th}$ cell

## Conclusion

The primary data type in LISP is the nestable list, i.e. the tree, so it makes sense that there is a shorthand language for accessing elements in trees. This is probably the driver that has caused locutions such as "*caddr*" to remain in the language. Perhaps the backronyms found in this paper will make such locutions easier to decipher (in this example: drop, drop, access). And now that we understand this access language, perhaps we might improve upon it.

(see http://www.textfiles.com/bitsavers/pdf/mit/computer_center/Coding_for_the_MIT-IBM_704_Computer_Oct57.pdf)

[3] "704 Electronic Data Processing Machine – Manual of Operation," 1954, 1955.
(see http://bitsavers.informatik.uni-stuttgart.de/pdf/ibm/704/24-6661-2_704_Manual_1955.pdf
http://www.cs.virginia.edu/brochure/images/manuals/IBM_704/IBM_704.html)

[4] http://www.iwriteiam.nl/HaCAR_CDR.html  chit chat with Steve Russel.

[5] Levin, Michael, "LISP 1.5 Programmer's Manual", 2$^{nd}$ edition, 1985.  MIT Press, p4.